\documentclass[11pt]{article}

\usepackage{acl}

\usepackage{times}
\usepackage{latexsym}
\usepackage[T1]{fontenc}
\usepackage[utf8]{inputenc}
\usepackage{microtype}
\usepackage{graphicx}
\usepackage{amsmath}
\usepackage{amssymb}
\usepackage{booktabs}
\usepackage{multirow}
\usepackage{xcolor}
\usepackage{tabularx}
\usepackage{subcaption}

\title{PeopleSearchBench: Evaluating AI-Powered People Search Platforms\\with Criteria-Grounded Verification}

\author{
\begin{tabular}{ccccccc}
Tianyu Shi$^{2}$ &
Wei Wang$^{1}$ &
Zequn Xie$^{1}$ &
Shuai Zhang$^{1}$ &
Boyang Xia$^{1}$ &
Chenyu Zeng$^{1}$ &
Qi Zhang$^{1}$
\\[2pt]
\multicolumn{7}{c}{
  \begin{tabular}{cccccc}
  Lynn Ai$^{1}$ &
  Yaqi Yu$^{1}$ &
  Kaiming Zhang$^{1}$ &
  Feiyue Tang$^{1}$ &
  Zhenyu Yu$^{1}$ &
  Lei Ding$^{3}$
  \end{tabular}
}
\end{tabular}
\\[6pt]
$^{1}$LESSIE Team
\\
$^{2}$McGill University, Montreal, QC, Canada
\\
$^{3}$Department of Statistics, University of Manitoba, Winnipeg, MB, Canada
\\[2pt]
\texttt{tianyu.shi3@mcgill.ca}\qquad
\texttt{shane@lessie.ai}
\qquad
\texttt{lei.ding@umanitoba.ca}
}

\begin{document}
\maketitle

\begin{abstract}
AI-powered people search platforms are increasingly deployed for recruiting, sales prospecting, and professional networking, yet no standardized benchmark exists for their rigorous evaluation. We present \textsc{PeopleSearchBench}, an open-source benchmark comprising 119 multilingual queries across four scenarios: corporate recruiting, B2B sales prospecting, expert search, and influencer discovery. A central contribution is \textit{Criteria-Grounded Verification}, an evaluation methodology that decomposes each query into explicit, independently checkable criteria and verifies each returned individual via live web search, producing factual relevance judgments rather than subjective LLM-as-judge scores (Cohen's $\kappa = 0.84$ with human annotators). We evaluate four architecturally diverse platforms along three complementary dimensions---Relevance Precision, Effective Coverage, and Information Utility---and find that multi-source search agents significantly outperform single-domain systems, particularly in influencer discovery where the performance gap is largest. Platform rankings are robust across ablations on scoring thresholds, dimension weights, and judge models. All code, queries, and evaluation prompts are publicly available.\footnote{Repository: \url{https://github.com/LessieAI/people-search-bench}}
\end{abstract}

\section{Introduction}

Finding the right person is a routine but high-stakes form of search. A recruiter may need senior backend engineers in London with microservices experience; a sales team may look for innovation leaders at large European enterprises; a researcher may need domain experts with specific publication histories; and a marketing team may seek creators with substantial audience reach in a particular technical community. In each case, the desired output is not a document or a single answer. It is a ranked set of people, each of whom must satisfy several constraints and provide enough verified information to support a downstream decision.

AI-powered people-search platforms are increasingly used to automate these workflows. Modern systems can query professional databases, search the open web, synthesize evidence from social platforms, and produce profile-level summaries. Yet evaluation has not kept pace with deployment. In the absence of a shared benchmark, it is difficult to tell whether a system has found genuinely qualified individuals, or whether it has returned plausible-looking profiles that match only surface keywords, omit important constraints, or lack actionable evidence.

Fundamentally, people search creates an evaluation problem that differs from standard information retrieval. Retrieval benchmarks such as BEIR and MTEB evaluate documents or passages as the primary result units~\citep{thakur2021beir,muennighoff2023mteb}. People search, by contrast, returns real individuals whose relevant attributes are distributed across multiple sources and may change over time. A person's current employer, role, location, seniority, publication record, or audience size cannot be treated as static text attached to a single result. These attributes must be verified against external evidence, and the evidence must be specific enough to explain why the person satisfies the search intent.

This requirement also limits the usefulness of holistic LLM-as-judge evaluation. LLM judges can approximate human preferences in many settings~\citep{zheng2023judging}, but a single subjective relevance score is a poor substitute for checking whether a named person currently holds a particular role, works at a specified company, or meets a quantitative constraint. Recent factuality methods decompose generated text into checkable claims~\citep{min2023factscore,wei2024safe}. People search calls for a complementary decomposition: the query itself should be converted into explicit criteria, and each returned person should be verified against those criteria using external evidence.

We present \textsc{PeopleSearchBench}, an open benchmark for evaluating AI-powered people-search platforms. The benchmark contains 119 multilingual queries spanning four practical scenarios: recruiting, B2B sales prospecting, expert search, and influencer/KOL discovery. Each query is decomposed into independently checkable criteria, and each returned person is evaluated through \textit{Criteria-Grounded Verification}, a decompose-then-verify pipeline that uses live web search to produce factual relevance judgments rather than relying on holistic impressions. The benchmark further evaluates whether systems return enough qualified people and whether the resulting profiles are usable for downstream action.

We make three contributions. First, we introduce Criteria-Grounded Verification, which extracts explicit criteria from a people-search query and verifies each returned individual against those criteria with external evidence, achieving $\kappa = 0.84$ agreement with human annotators. Second, we define a multi-dimensional evaluation framework that combines Relevance Precision, Effective Coverage, and Information Utility, capturing not only whether the top results are relevant but also whether the platform finds enough qualified people and provides actionable profile information. Third, we benchmark four architecturally diverse platforms across 119 queries, with confidence intervals, ablation studies, and error analysis. Our results show that performance varies substantially across scenarios, and that systems drawing on more diverse information sources are more robust across heterogeneous people-search tasks.

\section{Related Work}

\paragraph{Information retrieval benchmarks.}
TREC~\citep{voorhees2005trec} established the standard for IR evaluation with test collections and pooled relevance judgments.
BEIR~\citep{thakur2021beir} broadened the scope to 18 heterogeneous datasets for zero-shot retrieval, MTEB~\citep{muennighoff2023mteb} extended this to embedding models across diverse tasks, and BERGEN~\citep{rau2024bergen} benchmarked full retrieval-augmented generation pipelines.
Beyond text-only retrieval, cross-modal research has examined how video representations and alignment mechanisms affect retrieval.
HVD~\citep{xie2026hvd} introduces human-inspired coarse-to-fine visual selection and compression for text--video retrieval, while DPDV~\citep{xie2026dpdv} addresses cross-modal information asymmetry through dual-pathway and dual-view representation learning.
These benchmarks and methods evaluate documents, passages, or media items as self-contained retrieval units.
People search differs fundamentally: each result is a real individual with multiple attributes that must be independently verified.

\paragraph{LLM-based and factual evaluation.}
\citet{zheng2023judging} demonstrated that LLM judges can approximate human preferences, with subsequent work addressing positional bias~\citep{wang2024fair}, multi-dimensional rubrics~\citep{kim2024prometheus}, and score calibration~\citep{ye2024flask}.
A parallel line of work decomposes outputs into atomic claims for factual verification: FActScore~\citep{min2023factscore} checks LLM-generated biographies against Wikipedia, and SAFE~\citep{wei2024safe} extends this with web search.
Our Criteria-Grounded Verification shares the decompose-then-verify philosophy but differs in two key ways: we decompose the \textit{query} (not the output) into verifiable criteria, and we verify \textit{real individuals} rather than generated text, requiring live web evidence about current employment, location, and role.

\paragraph{Entity-centric and expert search.}
Entity retrieval from knowledge bases~\citep{hasibi2017dbpedia,balog2018entity} and expertise retrieval within closed corpora~\citep{balog2012expertise} assume fixed entity collections with known attributes.
\citet{geyik2018talent} described LinkedIn's talent search, evaluated using platform-specific engagement signals.
The term \textit{person retrieval} is also used in vision-language research to describe retrieving images of a target person from a fixed gallery using natural-language descriptions.
Recent work addresses noisy image--text correspondences through uncertainty-aware learning~\citep{xiedynamic} and uses multi-turn, chat-driven generation and interaction to enrich supervision and refine textual queries~\citep{xie2025chat}.
Despite sharing a person-centric matching objective, these methods retrieve visual instances from closed datasets rather than identify and verify professional profiles on the open web.
Consequently, these settings do not support cross-platform comparison over open-web results, nor do they provide evaluation protocols for autonomous, LLM-powered people-search agents.

\paragraph{Agentic AI evaluation.}
Recent benchmarks evaluate agents on software engineering~\citep{jimenez2024swe}, and general task completion~\citep{liu2024agentbench}.
These benchmarks primarily assess whether an agent completes a single well-defined goal.
Complementary work targets the accuracy--efficiency trade-off in web agents: SlimSearcher uses adaptive reward gating to encourage efficient search behavior~\citep{xie2026slimsearcher}, while WebClipper applies graph-based trajectory pruning to remove redundant interactions~\citep{wang2026webclipper}.
Although these approaches extend agent evaluation to interaction cost and trajectory efficiency, they do not assess the quality of a returned set of real-world individuals.
People search requires evaluating set-level relevance, per-result factual verification, and information quality jointly, a combination not addressed by existing agent benchmarks.

\section{\textsc{PeopleSearchBench}}

\begin{figure*}[t]
\centering
\includegraphics[width=1 \textwidth]{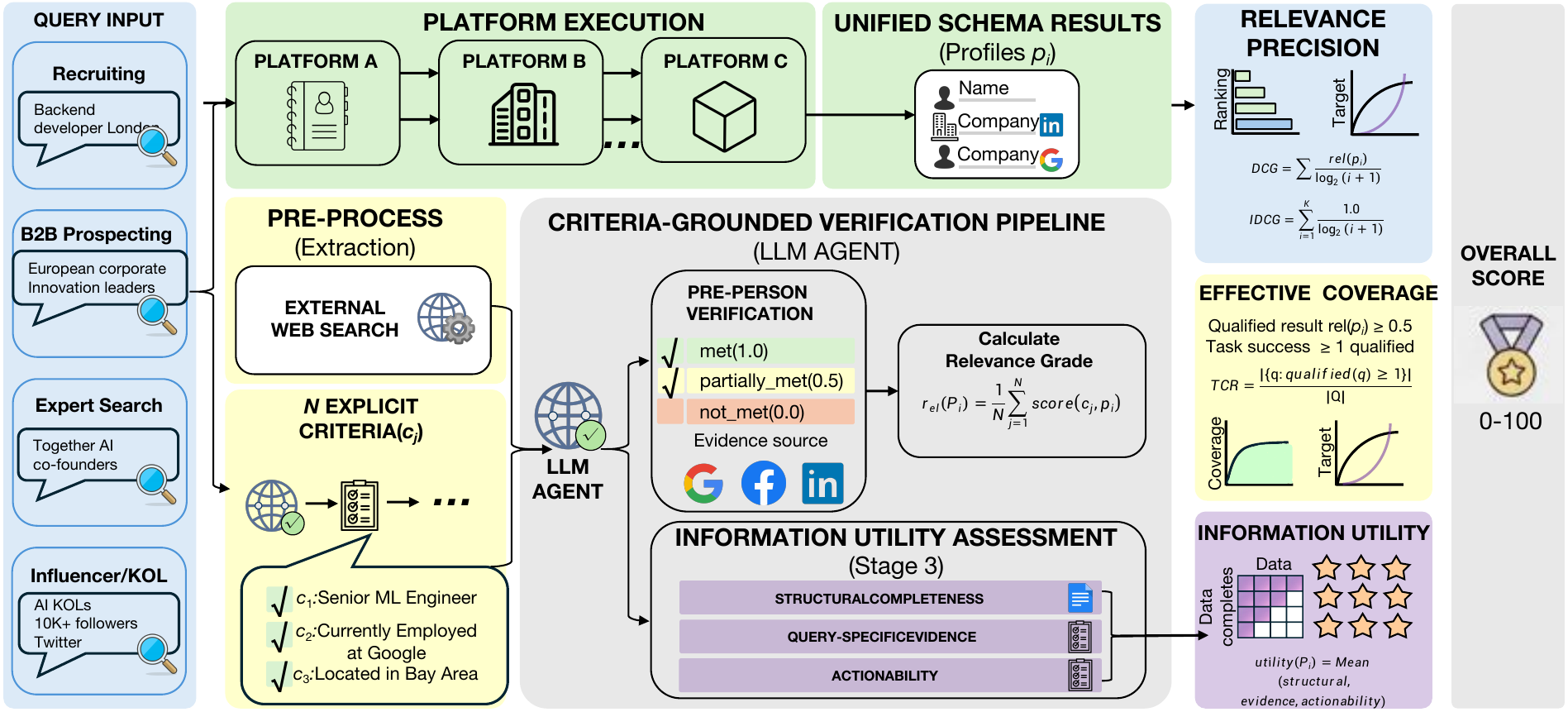}
\caption{Overview of the \textsc{PeopleSearchBench} evaluation pipeline. Queries are executed across platforms, results are normalized to a unified schema, and each person is verified against extracted criteria via web search. Scores along three dimensions are averaged into an overall score.}
\label{fig:pipeline}
\end{figure*}

Unlike document retrieval, people search returns real individuals whose attributes must be verified against external evidence and whose profiles must be actionable for downstream tasks. Our pipeline (Figure~\ref{fig:pipeline}) captures these requirements: queries from four scenarios are executed across platforms, results are normalized to a unified schema, and each person is verified against criteria extracted from the query via live web search. We describe the query dataset (\S\ref{sec:queries}), the verification pipeline (\S\ref{sec:pipeline}), and the scoring metrics (\S\ref{sec:dimensions}) below.

\subsection{Query Dataset}
\label{sec:queries}

The benchmark comprises 119 queries designed to reflect the actual needs of practitioners across four commercially important scenarios (Table~\ref{tab:query_dist}).

\textbf{Recruiting} (30 queries) targets candidates with specific combinations of skills, experience levels, and geographic preferences .
\textbf{B2B Prospecting} (32 queries) targets decision-makers at potential customer companies .
\textbf{Expert/Deterministic Search} (28 queries) seeks specific domain experts or has verifiable correct answers (e.g., ``Find all co-founders of [Company]'' or ``List all research scientists at [Organization]''). This category is particularly useful for validating factual accuracy since ground-truth answers can be established.
\textbf{Influencer/KOL Discovery} (29 queries) targets content creators and thought leaders in specific domains . This scenario tends to produce the largest performance differences, as influencer data is scattered across social platforms rather than concentrated in professional databases.

The query set is intentionally multilingual, covering English, Portuguese, Spanish, and Dutch, to reflect the global nature of modern people search. Each query contains on average 2.7 independently verifiable criteria, ranging from simple role-location pairs to complex multi-constraint specifications.

\begin{table}[t]
\centering
\small
\begin{tabular}{lcccc}
\toprule
\textbf{Category} & \textbf{N} & \textbf{Langs} & \textbf{Avg.~Crit.} & \textbf{Det.} \\
\midrule
Recruiting & 30 & EN, PT, ES & 3.2$\pm$1.1 & 0\% \\
B2B Prospect. & 32 & EN, ES & 2.8$\pm$0.9 & 0\% \\
Expert/Det. & 28 & EN & 2.1$\pm$0.7 & 100\% \\
Influencer & 29 & EN, NL, ES & 2.6$\pm$1.0 & 0\% \\
\midrule
Total & 119 & 4 & 2.7$\pm$1.0 & 23.5\% \\
\bottomrule
\end{tabular}
\caption{Query distribution. \textit{Avg.~Crit.}: mean extracted criteria per query. \textit{Det.}: percentage of queries with deterministic answers.}
\label{tab:query_dist}
\end{table}

\subsection{Criteria-Grounded Verification}
\label{sec:pipeline}

Our evaluation pipeline decomposes each query into explicit, verifiable factual checks grounded in external evidence, rather than assigning holistic quality scores as in standard LLM-as-judge methods (Table~\ref{tab:cgv_comparison}). This decompose-then-verify approach is conceptually related to FActScore~\citep{min2023factscore}, but differs in that we decompose the \textit{search query} into criteria and verify each returned \textit{person} against them using live web search, rather than decomposing generated text into atomic claims. The pipeline proceeds in three stages.

\begin{table}[t]
\centering
\small
\setlength{\tabcolsep}{4pt}
\begin{tabular}{lp{2.2cm}p{2.5cm}}
\toprule
\textbf{Aspect} & \textbf{LLM-as-Judge} & \textbf{Ours} \\
\midrule
Judgment & Subjective quality score (0--10) & Factual yes/no per criterion \\
Evidence & Parametric knowledge & External web search verification \\
Reproducibility & Low (prompt-sensitive) & High (criteria are explicit) \\
Bias risk & High (style, length bias) & Low (binary factual checks) \\
\bottomrule
\end{tabular}
\caption{Criteria-Grounded Verification vs.\ traditional holistic LLM-as-judge approaches.}
\label{tab:cgv_comparison}
\end{table}

\paragraph{Stage 1: Criteria Extraction.}
For each query, an LLM extracts $N$ explicit, independently checkable conditions from the stated search intent. For example:

\vspace{0.3em}
\noindent\textit{Query:} ``Find senior ML engineers at Google in Bay Area''\\
$\rightarrow$ $c_1$: Role is Senior ML Engineer or equivalent\\
$\rightarrow$ $c_2$: Currently employed at Google\\
$\rightarrow$ $c_3$: Located in San Francisco Bay Area
\vspace{0.3em}

\noindent Each criterion targets a single verifiable attribute, so the basis for every relevance judgment can be traced back to a specific factual check. To validate the stability of this step, we run the extraction prompt five times on 30 randomly selected queries with temperature 0.7 and find 94.7\% semantic equivalence across runs (\S\ref{sec:validation}).

\paragraph{Stage 2: Per-Person Verification.}
Each person returned by the platform is verified against every extracted criterion using live web search via the Tavily Search API with advanced depth settings. Each criterion receives one of three judgments:

\begin{itemize}
\setlength\itemsep{0em}
\item \textit{met} (1.0): the criterion is fully satisfied with external evidence
\item \textit{partially met} (0.5): the criterion is partially satisfied
\item \textit{not met} (0.0): no supporting or contradicting evidence
\end{itemize}

\noindent The person's relevance grade is the mean of the individual criterion scores:
\begin{equation}
\text{rel}(p_i) = \frac{1}{N}\sum_{j=1}^{N}\text{score}(c_j, p_i)
\label{eq:rel}
\end{equation}

\noindent This formulation treats all criteria as equally important. While weighted criteria could capture differences in constraint difficulty, equal weighting avoids introducing additional hyperparameters and is validated by our ablation studies (\S\ref{sec:ablation}).

\paragraph{Stage 3: Information Utility Assessment.}
The verification agent simultaneously scores each returned profile on whether it contains sufficient information---contact details, match explanations, source links---for downstream action (\S\ref{sec:info_util}).

\subsection{Evaluation Dimensions}
\label{sec:dimensions}

Each platform is scored on three independently computed dimensions, all scaled to the 0--100 range, then combined via equal-weight averaging to produce an overall score.

\paragraph{Relevance Precision (Padded nDCG@$K$).}
Relevance Precision measures whether the returned people match the query and are correctly ranked. Given relevance grades $\text{rel}(p_1), \ldots, \text{rel}(p_K)$ for the top-$K$ results, Discounted Cumulative Gain is:
\begin{equation}
\text{DCG@}K = \sum_{i=1}^{K}\frac{\text{rel}(p_i)}{\log_2(i+1)}
\label{eq:dcg}
\end{equation}

\noindent Unlike standard nDCG, which normalizes against the best possible ordering of the \textit{returned} results, we use a padded ideal that always assumes $K$ perfectly relevant results:
\begin{equation}
\text{IDCG@}K = \sum_{i=1}^{K}\frac{1.0}{\log_2(i+1)}
\label{eq:idcg}
\end{equation}

\noindent This prevents platforms that return only a few perfect results from receiving inflated scores---in people search, a platform returning 3 relevant people should score lower than one returning 10. Relevance Precision is the mean padded nDCG@$K$ across all queries, scaled to 0--100:
\begin{equation}
\text{Rel.\ Precision} = \frac{1}{|Q|}\sum_{q \in Q}\frac{\text{DCG@}K(q)}{\text{IDCG@}K} \times 100
\label{eq:rel_prec}
\end{equation}

\noindent We set $K{=}10$ for all main results, as most queries return at least 10 candidates. Sensitivity to this choice is examined in \S\ref{sec:ablation}.

\paragraph{Effective Coverage.}
Effective Coverage measures how many qualified people the platform finds per query, combining task completion with result volume. We define:

\begin{itemize}
\setlength\itemsep{0em}
\item \textit{Qualified result}: a person with $\text{rel}(p_i) \geq 0.5$, meaning they satisfy at least half of the extracted criteria.
\item \textit{Task success}: a query for which the platform returns at least one qualified result.
\end{itemize}

\noindent The coverage score is:
\begin{equation}
\text{Coverage} = \text{TCR} \times \frac{1}{|Q|}\sum_{q \in Q}\min\!\left(\frac{\text{qual}(q)}{K}, 1.0\right) \times 100
\label{eq:cov}
\end{equation}

\noindent where $\text{TCR} = |\{q : \text{qual}(q) \geq 1\}| / |Q|$ is the task completion rate and $K{=}10$ is the target number of results.

\paragraph{Information Utility.}
\label{sec:info_util}
Information Utility measures whether the returned data is sufficiently complete and structured for downstream action without further manual research. Each result is scored on three equally weighted sub-dimensions:

\begin{enumerate}
\setlength\itemsep{0em}
\item \textit{Structural completeness}: the richness of the profile data, including name, title, company, contact, work history, and education.
\item \textit{Query-specific evidence}: whether the result includes explanations for why the person matches each criterion and provides source URLs for verification.
\item \textit{Actionability}: whether the user can take next steps (contact, shortlisting, outreach) based on the provided data alone.
\end{enumerate}

\noindent Each sub-dimension is scored on a 0.0--1.0 scale:

\begin{equation}
\text{utility}(p_i) = \frac{1}{3}\big(\text{struct} + \text{evid} + \text{action}\big)
\label{eq:util_person}
\end{equation}

\begin{equation}
\text{Info.\ Util} = \frac{100}{|Q|} \sum_{q \in Q} \frac{1}{|P_q|} \sum_{p_i \in P_q} \text{utility}(p_i)
\label{eq:util}
\end{equation}
\noindent where $P_q$ is the set of evaluated persons for query $q$.

\paragraph{Overall Score.}
The three dimensions are combined via equal-weight averaging:
\begin{equation}
\text{Overall} = \frac{\text{Rel.\ Prec.} + \text{Eff.\ Cov.} + \text{Info.\ Util.}}{3}
\label{eq:overall}
\end{equation}

\noindent We follow the Multi-Criteria Decision Analysis principle that equal weights perform comparably to optimized weights in most multi-attribute decision problems~\citep{dawes1974linear}. Our ablation study (\S\ref{sec:ablation}) confirms that rankings are invariant to weight perturbation.

\section{Experiments}

\subsection{Setup}

\paragraph{Platforms.}
We evaluate four platforms representing diverse architectural approaches to people search (Table~\ref{tab:platforms}).
\textbf{Lessie} is a specialized AI search agent that autonomously searches across professional networks, social platforms, academic databases, and public registries.
\textbf{Exa} is a structured search API that returns entity results from a proprietary database.
\textbf{Juicebox} (PeopleGPT) is an AI recruiting platform with access to over 800 million professional profiles from 60+ sources.
\textbf{Claude Code} is a general-purpose AI coding agent (Claude Sonnet 4.6) that produces text-based search reports via web search.
All platforms are evaluated on up to 15 results per query.

\paragraph{Verification configuration.}
The verification pipeline uses Gemini 3 Flash Preview (via OpenRouter) for all LLM judgments and the Tavily Search API (advanced depth) for web-based fact-checking. The same model and configuration are applied identically to all platforms, and the verification agent receives no information about which platform produced each result.

\paragraph{Temporal control.}
All evaluations were conducted within a single week (January 15--22, 2025), with each platform evaluated on the same day using identical query ordering. Platform versions were recorded: Lessie v2.1.0 (web interface), Exa API v1 (entity search endpoint), Juicebox PeopleGPT v3.2 (web interface), and Claude Code (claude-sonnet-4-6-20250101, via API).

\paragraph{Statistical methodology.}
We report 95\% confidence intervals estimated via bootstrap resampling (1000 iterations) and assess pairwise significance using paired bootstrap tests~\citep{efron1994bootstrap}.

\begin{table}[t]
\centering
\small
\setlength{\tabcolsep}{4pt}
\begin{tabular}{llp{3.5cm}}
\toprule
\textbf{Platform} & \textbf{Type} & \textbf{Data Sources} \\
\midrule
Lessie & Specialized agent & Web, social, academic, public registries \\
Exa & Search API & Proprietary entity database \\
Juicebox & Recruiting platform & 800M+ profiles, 60+ sources \\
Claude Code & General AI agent & Web search \\
\bottomrule
\end{tabular}
\caption{Characteristics of the four evaluated platforms.}
\label{tab:platforms}
\end{table}

\subsection{Main Results}

\begin{figure}[t]
\centering
\includegraphics[width=\linewidth]{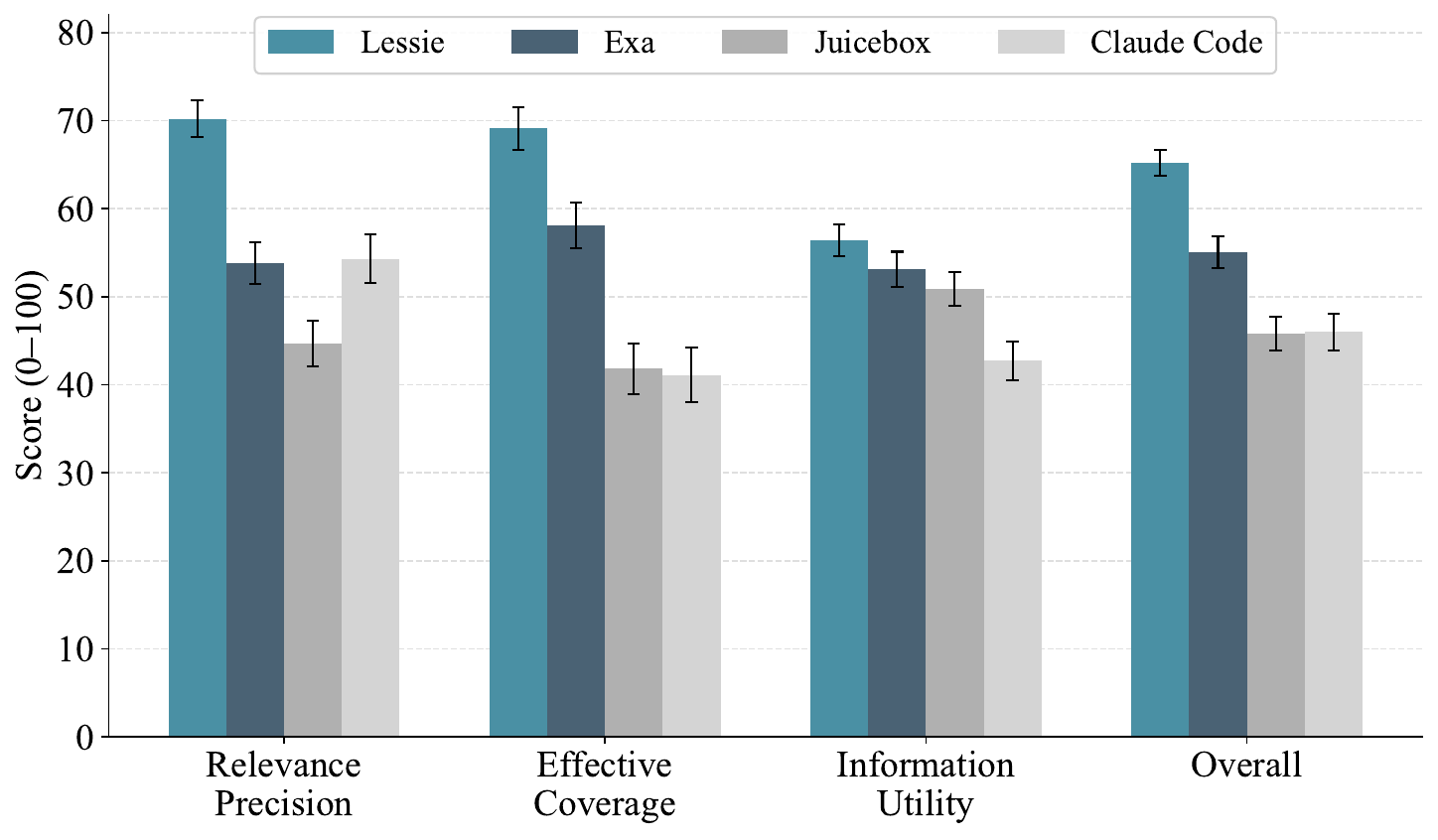}
\caption{Overall results decomposed by dimension with 95\% confidence intervals.}
\label{fig:overall_comparison}
\end{figure}

Lessie achieves the highest overall score (65.2$\pm$1.5), leading across all three evaluation dimensions with an 18.5\% margin over the second-ranked system (Figure~\ref{fig:overall_comparison}). Exa ranks second (55.0$\pm$1.8), followed by Claude Code (46.0$\pm$2.1) and Juicebox (45.8$\pm$1.9). All pairwise differences between the first- and second-ranked platforms are statistically significant ($p < 0.05$, paired bootstrap).

The per-dimension breakdown reveals distinct architectural tradeoffs. Exa achieves strong Effective Coverage (58.1$\pm$2.6) due to its high task completion rate (96.6$\pm$1.8\%) and consistent result counts, but its Relevance Precision (53.8$\pm$2.4) trails Lessie by 16.4 points, suggesting difficulty with complex multi-constraint queries. Claude Code achieves moderate Relevance Precision (54.3$\pm$2.8) through general-purpose web search but the lowest Information Utility (42.7$\pm$2.2), as its text-based reports typically lack structured contact information and per-criterion match explanations. Juicebox has the lowest Relevance Precision (44.7$\pm$2.6), indicating that its recruiting-focused database design generalizes poorly to non-recruiting queries, though its rich LinkedIn-style profile fields yield moderate Information Utility (50.9$\pm$1.9).

\begin{figure}[t]
\centering
\includegraphics[width=\linewidth]{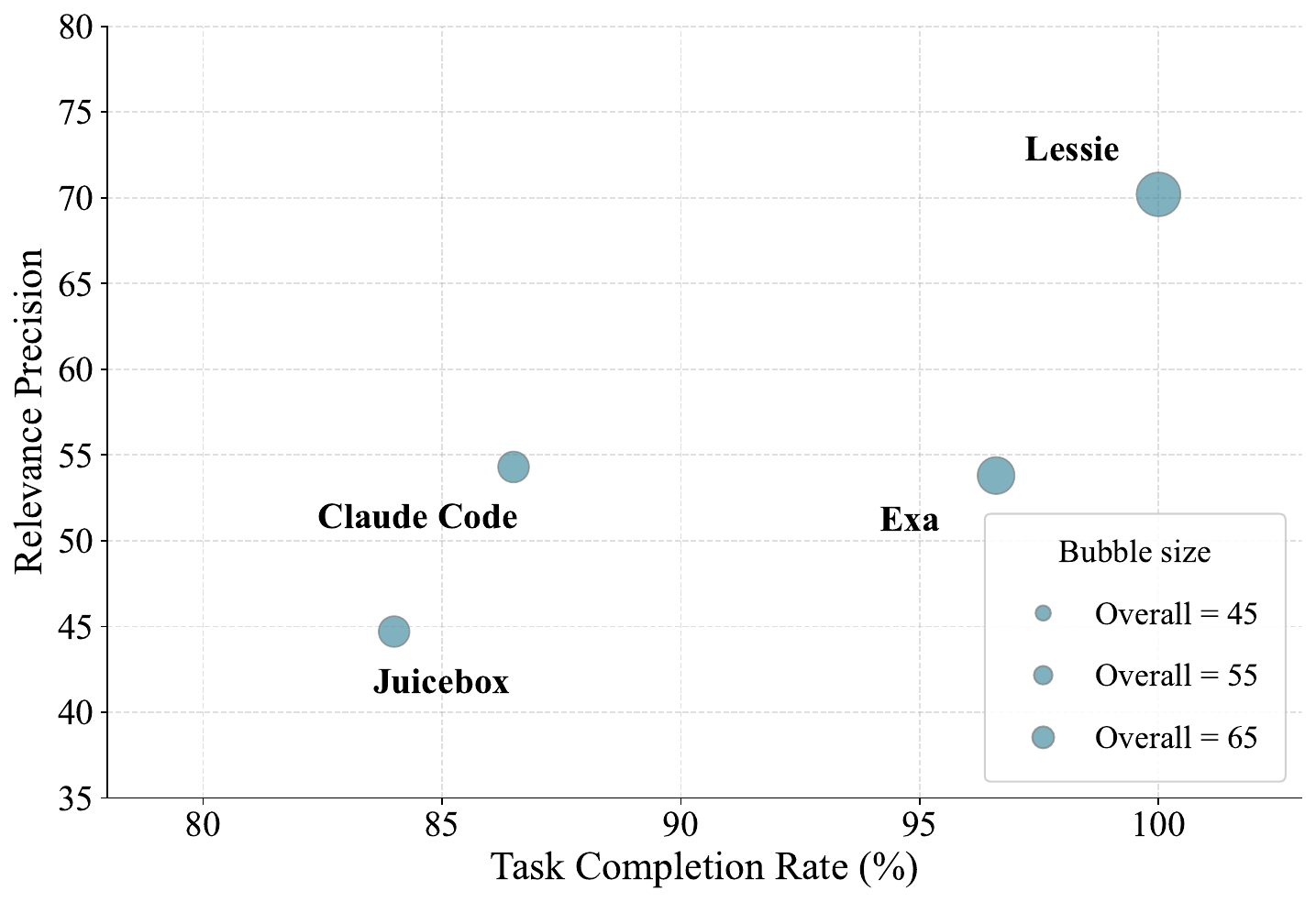}
\caption{Task completion rate vs.\ Relevance Precision. Bubble size indicates overall score.}
\label{fig:completion_vs_relevance}
\end{figure}

Lessie is notably the only platform achieving both 100\% task completion and the highest relevance, occupying the upper-right quadrant of the completion--relevance space (Figure~\ref{fig:completion_vs_relevance}). The remaining platforms trade off between these two dimensions: Exa achieves high completion but moderate relevance, while Juicebox and Claude Code cluster in the lower-left with both lower completion and lower relevance.

\section{Conclusion}

We presented \textsc{PeopleSearchBench}, a benchmark with Criteria-Grounded Verification for evaluating AI-powered people search platforms.
The benchmark evaluates systems along three complementary dimensions---Relevance Precision, Effective Coverage, and Information Utility---using a decompose-then-verify pipeline that produces factual, reproducible judgments with high human agreement ($\kappa = 0.84$).
Evaluation of four platforms---Lessie, Exa, Juicebox, and Claude Code---on 119 queries across four scenarios reveals clear architectural tradeoffs, with data source diversity playing a key role in cross-scenario robustness.
The benchmark's design choices---padded nDCG, qualified-result thresholds, equal dimension weighting---are validated through systematic ablation studies showing stable rankings.
We release all code, queries, and evaluation prompts to support reproducible comparison as AI-powered people search continues to evolve.

\section*{Limitations}

Our benchmark has several limitations.
(1)~The 119-query set, while balanced across four scenarios, does not cover every people-search use case (e.g., academic collaborator search, angel investor identification).
(2)~Web verification depends on publicly indexed information; individuals with limited online presence may be systematically under-evaluated, introducing a bias toward publicly visible professionals.
(3)~We evaluate up to 15 results per query; platforms capable of returning more are only assessed on their top 15.
(4)~Platform capabilities evolve quickly; our results reflect a single evaluation window in January 2025 and may not generalize to later versions.
(5)~The Information Utility dimension rewards platforms providing per-result match explanations, which may favor architectures with built-in verification pipelines.
(6)~While we test multiple judge models and find consistent rankings, the primary pipeline uses a single LLM verifier, and all tested judges share similar architectural foundations.

\section*{Acknowledgments}
Lei Ding was supported by the Natural Sciences and Engineering Research Council of Canada (NSERC) Discovery Grant (RGPIN-2026-06465).

\section*{Ethics Statement}

All queries target publicly available professional information; we do not scrape private data.
The benchmark publishes aggregated platform scores, not individual personal records.
Human annotators were compensated above local minimum wage and provided informed consent.
We recognize that people-search technology can be misused and encourage adopters to pair this benchmark with responsible data-handling policies.

\bibliography{references}

@inproceedings{thakur2021beir,
  title={{BEIR}: A Heterogeneous Benchmark for Zero-shot Evaluation of Information Retrieval Models},
  author={Thakur, Nandan and Reimers, Nils and R{\"u}ckl{\'e}, Andreas and Srivastava, Abhishek and Gurevych, Iryna},
  booktitle={Thirty-fifth Conference on Neural Information Processing Systems Datasets and Benchmarks Track},
  year={2021}
}

@article{xiedynamic,
  title={Dynamic Uncertainty Learning with Noisy Correspondence for Text-based Person Retrieval},
  author={Xie, Zequn and Wang, Chuxin and Cai, Sihang and Feng, Fangming and luo, Guijin and Zhang, Qifei and Chen, Jingyuan and Jin, Tao},
  journal={ACM Transactions on Information Systems},
  publisher={ACM New York, NY}
}

@article{xie2026slimsearcher,
  title={SlimSearcher: Training Efficiency-Aware Web Agents via Adaptive Reward Gating},
  author={Xie, Zequn and Wang, Junjie and Yang, Dan and Feng, Jie and Shen, Yue and Wang, Jian and Gu, Jinjie},
  journal={arXiv preprint arXiv:2606.07074},
  year={2026}
}

@inproceedings{xie2026dpdv,
  title={DPDV: Dual-Pathway and Dual-View Representation Learning for Bridging Information Asymmetry in Text-Video Retrieval},
  author={Xie, Zequn and Liu, Xin and Feng, Fangming and Zhang, Boyun and Jin, Tao},
  booktitle={Proceedings of the 64th Annual Meeting of the Association for Computational Linguistics (Volume 1: Long Papers)},
  pages={5385--5394},
  year={2026}
}

@inproceedings{wang2026webclipper,
  title={WebClipper: Efficient Evolution of Web Agents with Graph-based Trajectory Pruning},
  author={Wang, Junjie and Xie, Zequn and Yang, Dan and Feng, Jie and Shen, Yue and Sun, Duolin and Long, Meixiu and Jiao, Yihan and Tan, Zhehao and Wang, Jian and others},
  booktitle={Proceedings of the 64th Annual Meeting of the Association for Computational Linguistics (Volume 1: Long Papers)},
  pages={21646--21666},
  year={2026}
}

@inproceedings{xie2026hvd,
  title     = {{HVD}: Human Vision-Driven Video Representation Learning
               for Text-Video Retrieval},
  author    = {Xie, Zequn and Liu, Xin and Zhang, Boyun and Lin, Yuxiao
               and Cai, Sihang and Jin, Tao},
  booktitle = {ICASSP 2026--2026 IEEE International Conference on
               Acoustics, Speech and Signal Processing (ICASSP)},
  pages     = {12992--12996},
  year      = {2026},
  publisher = {IEEE},
  doi       = {10.1109/ICASSP55912.2026.11464612}
}

@inproceedings{xie2025chat,
  title={Chat-driven text generation and interaction for person retrieval},
  author={Xie, Zequn and Wang, Chuxin and Wang, Yeqiang and Cai, Sihang and Wang, Shulei and Jin, Tao},
  booktitle={Proceedings of the 2025 Conference on Empirical Methods in Natural Language Processing},
  pages={5259--5270},
  year={2025}
}

@inproceedings{muennighoff2023mteb,
  title={{MTEB}: Massive Text Embedding Benchmark},
  author={Muennighoff, Niklas and Tazi, Nouamane and Magne, Lo{\"\i}c and Reimers, Nils},
  booktitle={Proceedings of the 17th Conference of the European Chapter of the Association for Computational Linguistics},
  pages={2014--2037},
  year={2023}
}

@inproceedings{zheng2023judging,
  title={Judging {LLM}-as-a-Judge with {MT-Bench} and Chatbot Arena},
  author={Zheng, Lianmin and Chiang, Wei-Lin and Sheng, Ying and Zhuang, Siyuan and Wu, Zhanghao and Zhuang, Yonghao and Lin, Zi and Li, Zhuohan and Li, Dacheng and Xing, Eric P and others},
  booktitle={Advances in Neural Information Processing Systems},
  volume={36},
  year={2023}
}

@inproceedings{min2023factscore,
  title={{FActScore}: Fine-grained Atomic Evaluation of Factual Precision in Long Form Text Generation},
  author={Min, Sewon and Krishna, Kalpesh and Lyu, Xinxi and Lewis, Mike and Yih, Wen-tau and Koh, Pang Wei and Iyyer, Mohit and Zettlemoyer, Luke and Hajishirzi, Hannaneh},
  booktitle={Proceedings of the 2023 Conference on Empirical Methods in Natural Language Processing},
  pages={12076--12100},
  year={2023}
}

@article{wei2024safe,
  title={{Long-form Factuality in Large Language Models}},
  author={Wei, Jerry and Yang, Chengrun and Song, Xinying and Lu, Yifeng and Hu, Nathan and Tran, Dustin and Peng, Daiyi and Liu, Ruibo and Huang, Da and Du, Cosmo and Le, Quoc V},
  journal={arXiv preprint arXiv:2403.18802},
  year={2024}
}

@inproceedings{rau2024bergen,
  title={{BERGEN}: A Benchmarking Library for Retrieval-Augmented Generation},
  author={Rau, David and D{\'e}jean, Herv{\'e} and Chirkova, Nadezhda and Formal, Thibault and Wang, Shuai and Clinchant, St{\'e}phane and Nikoulina, Vassilina},
  booktitle={Findings of the Association for Computational Linguistics: EMNLP 2024},
  pages={7640--7663},
  year={2024}
}

@book{voorhees2005trec,
  title={{TREC}: Experiment and Evaluation in Information Retrieval},
  author={Voorhees, Ellen M and Harman, Donna K},
  year={2005},
  publisher={MIT Press}
}

@inproceedings{wang2024fair,
  title={Large Language Models Are Not Fair Evaluators},
  author={Wang, Peiyi and Li, Lei and Chen, Liang and others},
  booktitle={Proceedings of the 62nd Annual Meeting of the Association for Computational Linguistics},
  pages={9440--9450},
  year={2024}
}

@inproceedings{kim2024prometheus,
  title={Prometheus 2: An Open Source Language Model Specialized in Evaluating Other Language Models},
  author={Kim, Seungone and Suk, Juyoung and Longpre, Shayne and others},
  booktitle={Proceedings of the 2024 Conference on Empirical Methods in Natural Language Processing},
  year={2024}
}

@inproceedings{ye2024flask,
  title={{FLASK}: Fine-grained Language Model Evaluation Based on Alignment Skill Sets},
  author={Ye, Seonghyeon and Kim, Doyoung and Kim, Sungdong and others},
  booktitle={International Conference on Learning Representations},
  year={2024}
}

@inproceedings{hasibi2017dbpedia,
  title={{DBpedia-Entity} v2: A Test Collection for Entity Search},
  author={Hasibi, Faegheh and Nikolaev, Fedor and Xiong, Chenyan and Balog, Krisztian and Bratsberg, Svein Erik and Kotov, Alexander and Callan, Jamie},
  booktitle={Proceedings of the 40th International ACM SIGIR Conference},
  pages={1265--1268},
  year={2017}
}

@book{balog2018entity,
  title={Entity-Oriented Search},
  author={Balog, Krisztian},
  volume={39},
  series={The Information Retrieval Series},
  year={2018},
  publisher={Springer}
}

@article{balog2012expertise,
  title={Expertise Retrieval},
  author={Balog, Krisztian and Fang, Yi and de Rijke, Maarten and Serdyukov, Pavel and Si, Luo},
  journal={Foundations and Trends in Information Retrieval},
  volume={6},
  number={2--3},
  pages={127--256},
  year={2012}
}

@inproceedings{geyik2018talent,
  title={Talent Search and Recommendation Systems at {LinkedIn}},
  author={Geyik, Sahin Cem and Guo, Qi and Hu, Bo and others},
  booktitle={Proceedings of the 41st International ACM SIGIR Conference},
  pages={1353--1354},
  year={2018}
}

@inproceedings{jimenez2024swe,
  title={{SWE}-bench: Can Language Models Resolve Real-World {GitHub} Issues?},
  author={Jimenez, Carlos E and Yang, John and Wettig, Alexander and others},
  booktitle={International Conference on Learning Representations},
  year={2024}
}

@inproceedings{liu2024agentbench,
  title={{AgentBench}: Evaluating {LLMs} as Agents},
  author={Liu, Xiao and Yu, Hao and Zhang, Hanchen and others},
  booktitle={International Conference on Learning Representations},
  year={2024}
}

@article{dawes1974linear,
  title={Linear Models in Decision Making},
  author={Dawes, Robyn M and Corrigan, Bernard},
  journal={Psychological Bulletin},
  volume={81},
  number={2},
  pages={95--106},
  year={1974}
}

@book{efron1994bootstrap,
  title={An Introduction to the Bootstrap},
  author={Efron, Bradley and Tibshirani, Robert J},
  year={1994},
  publisher={Chapman and Hall/CRC}
}

\appendix

\subsection{Scenario Breakdown}

Performance varies substantially across query scenarios (Table~\ref{tab:scenario_detail}).

\textbf{Recruiting} is the most competitive scenario. Juicebox achieves second place (65.7$\pm$2.9) with the highest Effective Coverage in this category (75.3$\pm$2.7), reflecting the strength of its large professional database. Lessie leads overall (68.2$\pm$2.8), primarily through superior Relevance Precision (74.8$\pm$2.6). All four platforms achieve at least 90\% task completion here, indicating that recruiting queries are well-served by existing data sources.

\textbf{B2B Prospecting} reveals wider gaps. Lessie's Relevance Precision (62.8$\pm$2.9) leads Exa (50.0$\pm$3.2) by 12.8 points, suggesting that multi-source data fusion is especially valuable when queries target decision-makers outside standard professional databases. Juicebox's task completion drops to 84.4$\pm$6.5\%, and Claude Code falls further to 75.0$\pm$7.7\%, as B2B targets often lack the structured profiles these platforms rely on.

\textbf{Expert/Deterministic} queries favor Lessie (70.4$\pm$2.4), which achieves its highest Relevance Precision (79.0$\pm$2.3) in this category. Claude Code performs comparatively well overall (57.0$\pm$3.1), as deterministic queries benefit from general-purpose web search---specific known individuals can be located through standard search engines. However, Claude Code's Information Utility remains low (38.5$\pm$3.4) because its text reports lack structured profile data. Juicebox struggles most in this category (44.2$\pm$3.6), with task completion at only 71.4$\pm$8.5\%, as many target individuals lack LinkedIn profiles.

\textbf{Influencer/KOL} exhibits the widest performance spread. Lessie's Relevance Precision (65.2$\pm$3.1) is 2.45$\times$ Juicebox's (26.6$\pm$4.0), and Juicebox's task completion falls to 79.3$\pm$7.6\%. Since influencer data is distributed across Instagram, Twitter/X, and YouTube rather than concentrated in professional databases, multi-source architectures hold a pronounced advantage. This scenario also produces the largest absolute gap in overall score: 31.2 points between Lessie and Juicebox.

\begin{figure}[t]
\centering
\includegraphics[width=\linewidth]{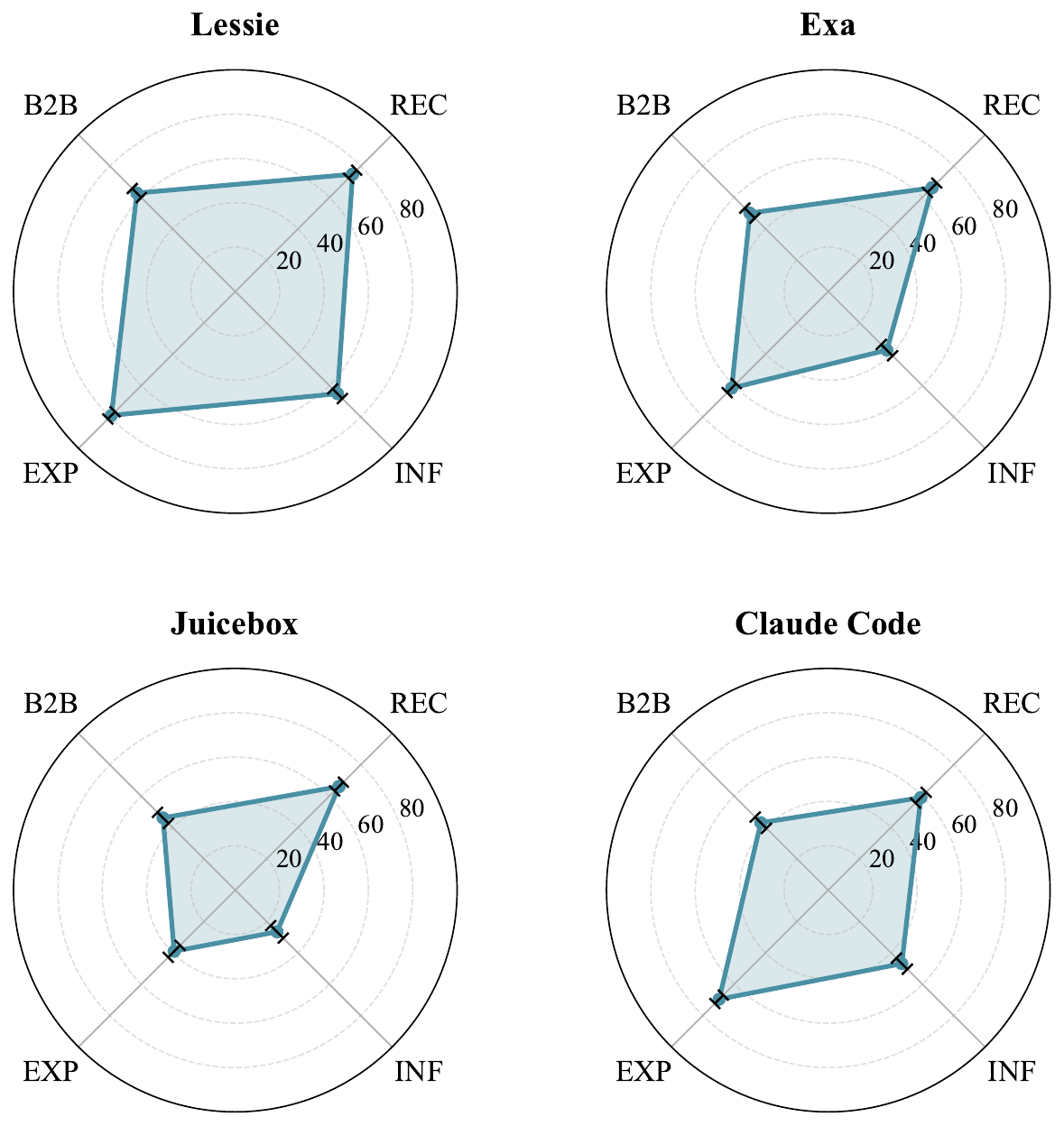}
\caption{Relevance Precision by scenario for each platform with 95\% bootstrap CIs. REC = Recruiting, B2B = B2B
  Prospecting, EXP = Expert/Deterministic, INF = Influencer/KOL.}
\label{fig:radar}
\end{figure}

\textbf{Cross-scenario consistency.}
Lessie is the only platform maintaining consistent Relevance Precision across all categories, with a range of 62.8--79.0 (coefficient of variation: 9.7\%), as visualized in Figure~\ref{fig:radar}. Other platforms exhibit significantly wider variance: Juicebox ranges from 26.6 to 66.1 (CV: 35.2\%), Exa from 37.4 to 66.2 (CV: 22.8\%), and Claude Code from 43.0 to 69.6 (CV: 19.1\%). These results suggest that multi-source architectures are structurally less sensitive to query type, whereas platforms built around a single data domain exhibit sharp performance degradation outside that domain.

\begin{table}[t]
\centering
\small
\setlength{\tabcolsep}{3pt}
\begin{tabular}{llcccc}
\toprule
\textbf{Scenario} & & \textbf{Lessie} & \textbf{Exa} & \textbf{Jbox} & \textbf{CC} \\
\midrule
\multirow{2}{*}{Recruiting}
  & Overall & \textbf{68.2}{\scriptsize$\pm$2.8} & 64.7{\scriptsize$\pm$3.1} & 65.7{\scriptsize$\pm$2.9} & 50.5{\scriptsize$\pm$3.5} \\
  & TCR (\%) & 100 & 100 & 100 & 90.0{\scriptsize$\pm$5.5} \\
\midrule
\multirow{2}{*}{B2B Prosp.}
  & Overall & \textbf{60.6}{\scriptsize$\pm$2.6} & 55.2{\scriptsize$\pm$2.9} & 51.4{\scriptsize$\pm$3.2} & 43.0{\scriptsize$\pm$3.4} \\
  & TCR (\%) & 100 & 100 & 84.4{\scriptsize$\pm$6.5} & 75.0{\scriptsize$\pm$7.7} \\
\midrule
\multirow{2}{*}{Expert/Det.}
  & Overall & \textbf{70.4}{\scriptsize$\pm$2.4} & 61.2{\scriptsize$\pm$2.8} & 44.2{\scriptsize$\pm$3.6} & 57.0{\scriptsize$\pm$3.1} \\
  & TCR (\%) & 100 & 96.4{\scriptsize$\pm$3.6} & 71.4{\scriptsize$\pm$8.5} & 100 \\
\midrule
\multirow{2}{*}{Influencer}
  & Overall & \textbf{62.3}{\scriptsize$\pm$3.0} & 41.6{\scriptsize$\pm$3.4} & 31.1{\scriptsize$\pm$3.8} & 43.2{\scriptsize$\pm$3.3} \\
  & TCR (\%) & 100 & 89.7{\scriptsize$\pm$5.7} & 79.3{\scriptsize$\pm$7.6} & 82.8{\scriptsize$\pm$7.0} \\
\bottomrule
\end{tabular}
\caption{Per-scenario overall scores and task completion rates (TCR) with 95\% bootstrap CIs. Jbox = Juicebox; CC = Claude Code.}
\label{tab:scenario_detail}
\end{table}

\section{Ablation Studies}
\label{sec:ablation}

We conduct ablation experiments on five design choices to assess the robustness of our benchmark; full tables are in Appendix~\ref{app:ablation}. The central finding is that platform rankings are stable across all tested configurations.

\begin{figure}[t]
\centering
\includegraphics[width=\linewidth]{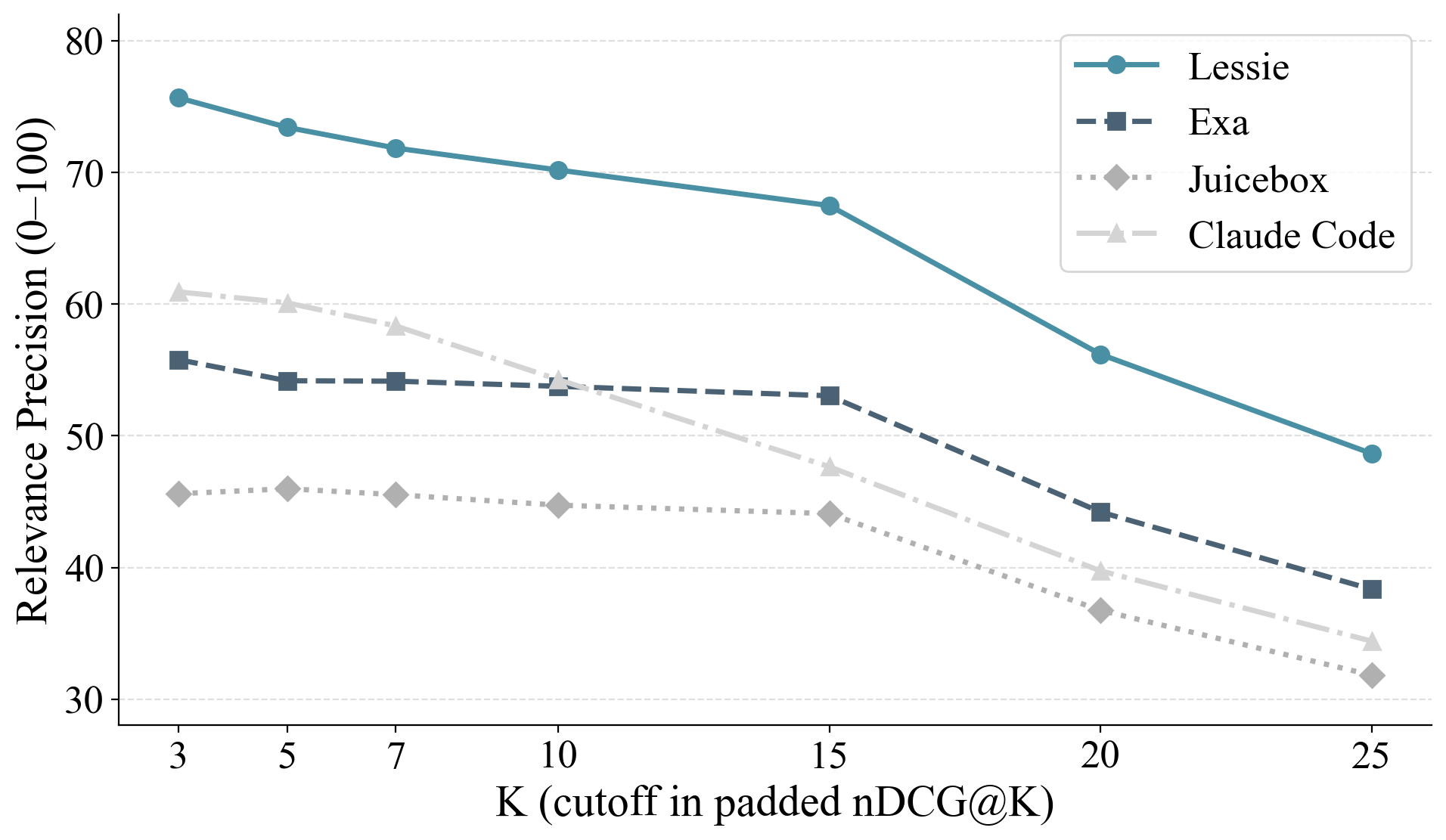}
\caption{Sensitivity of Relevance Precision to cutoff K across platforms.}
\label{fig:ndcg_vs_k}
\end{figure}

First, the scoring parameters. Varying the qualified-result threshold across $\text{rel}(p_i) \geq \{0.3, 0.5, 0.7\}$ preserves the top-two rankings (Lessie, Exa) at every level (Table~\ref{tab:threshold_sensitivity}). We also test the nDCG cutoff at $K \in \{3, 5, 7, 10, 15, 20, 25\}$. Lessie remains first across all values of $K$, and overall rankings are stable for $K \leq 10$. At $K{=}15$, Claude Code's score drops more steeply than others. At $K \geq 20$, all scores decline sharply since most platforms return fewer than 15 results (Table~\ref{tab:topk_sensitivity}, Figure~\ref{fig:ndcg_vs_k}). Finally, replacing the ternary criterion judgments with binary scoring lowers all Relevance Precision scores---e.g., Lessie from 70.2 to 68.4---but preserves all rankings (Table~\ref{tab:partial_ablation}).

Second, the aggregation scheme. We test five weighting schemes for combining the three dimensions, including weights optimized via grid search on a held-out set of 30 queries. Lessie ranks first under all five schemes (Table~\ref{tab:weight_sensitivity}). Removing Information Utility entirely and scoring on only Relevance Precision and Effective Coverage causes Juicebox (43.3) to fall below Claude Code (47.7), confirming that profile completeness is a meaningful quality axis not captured by the other two dimensions (Table~\ref{tab:utility_ablation}).

Taken together, these results indicate that the benchmark's conclusions are not sensitive to specific parameter choices or aggregation decisions.

\subsection{Pipeline Validation}
\label{sec:validation}

A core contribution of this work is the Criteria-Grounded Verification pipeline itself. We conduct three experiments to validate its reliability.

\paragraph{Human validation.}
Two trained annotators independently reviewed a stratified sample of 200 person--query pairs (50 per scenario), blinded to platform identity, with access to the same web search tools as the automated pipeline. Inter-annotator agreement is substantial: criterion match $\kappa = 0.87$ (95\% CI: 0.83--0.91); qualified status $\kappa = 0.91$ (0.87--0.95). Agreement between the LLM verifier and human consensus reaches $\kappa = 0.84$ (0.79--0.89) on criterion match and $\kappa = 0.88$ (0.83--0.93) on qualified status (Table~\ref{tab:human_val}). Among the 26 criterion-level disagreements, 18 (69\%) involved the LLM being more conservative than humans on ``partially met'' judgments, and 8 (31\%) involved the LLM finding evidence that humans missed. This pattern indicates that the pipeline is slightly conservative but not systematically biased toward any platform.

\paragraph{Criteria extraction stability.}
We run the extraction prompt five times on 30 randomly selected queries at temperature 0.7. The mean number of criteria per query is 2.73 (SD = 0.41), and semantic equivalence across runs---assessed by an independent LLM---reaches 94.7\%. Exact string match is 78.0\%, which serves as a lower bound since paraphrasing is acceptable.

\paragraph{Judge model sensitivity.}
We tested three alternative judge models on 200 person--query pairs. All show high agreement with the primary Gemini model: GPT-4o ($\kappa = 0.87$), Claude 3.5 Sonnet ($\kappa = 0.85$), and GPT-4o-mini ($\kappa = 0.79$). Platform rankings remain identical across all judges, indicating that the benchmark's conclusions are not artifacts of the specific judge model.

\begin{table}[t]
\centering
\small
\begin{tabular}{lccc}
\toprule
\textbf{Metric} & \textbf{Agree.} & $\boldsymbol{\kappa}$ & \textbf{95\% CI} \\
\midrule
Crit.\ match (3-level) & 86.5\% & 0.84 & [0.79, 0.89] \\
Qualified (binary) & 93.0\% & 0.88 & [0.83, 0.93] \\
Rel.\ grade (cont.) & --- & $r{=}0.89$ & [0.85, 0.92] \\
\bottomrule
\end{tabular}
\caption{LLM verifier vs.\ human consensus on 200 person--query pairs.}
\label{tab:human_val}
\end{table}

\section{Error Analysis}

We manually reviewed all queries with at least one error and categorized failures into four types (Table~\ref{tab:errors}). Per-scenario breakdowns and illustrative case studies are in Appendix~\ref{app:error}.

False positives are most prevalent in Juicebox (24.6\%), which frequently matches surface-level keywords---e.g., ``product manager'' + ``Singapore'' + ``finance''---without satisfying semantic constraints like ``fintech startup.'' Incomplete profiles dominate Claude Code's errors (31.2\%), as its text reports lack structured contact data and source URLs. False negatives and task failures concentrate in Juicebox outside its recruiting domain, reaching a 41.4\% false-negative rate in Influencer/KOL queries where content creators rarely maintain professional database profiles. Lessie maintains 0\% false-negative and task-failure rates across all scenarios.

The Influencer/KOL scenario exhibits the highest error rates across all platforms, confirming that influencer discovery---where relevant individuals are dispersed across heterogeneous social platforms---remains the most challenging category.

\begin{table}[t]
\centering
\small
\setlength{\tabcolsep}{3.5pt}
\begin{tabular}{lcccc}
\toprule
\textbf{Error Type} & \textbf{Lessie} & \textbf{Exa} & \textbf{Jbox} & \textbf{CC} \\
\midrule
False positive & 8.2 & 18.4 & 24.6 & 16.8 \\
False negative & 0 & 3.4 & 16.0 & 13.5 \\
Incomplete profile & 12.4 & 14.2 & 8.6 & 31.2 \\
Task failure & 0 & 3.4 & 16.0 & 13.5 \\
\bottomrule
\end{tabular}
\caption{Error rates (\%) by platform. Jbox = Juicebox; CC = Claude Code.}
\label{tab:errors}
\end{table}

\section{Ablation Details}
\label{app:ablation}

This appendix provides the full tables for the ablation studies summarized in \S\ref{sec:ablation}.

\paragraph{Qualified threshold sensitivity.}
We test three thresholds for defining a qualified result: $\text{rel}(p_i) \geq 0.3$, $0.5$, and $0.7$. Rankings are reported for both Effective Coverage and Overall score.

\begin{table}[h]
\centering
\small
\setlength{\tabcolsep}{3pt}
\begin{tabular}{lcccccc}
\toprule
& \multicolumn{3}{c}{\textbf{Cov.\ Rank}} & \multicolumn{3}{c}{\textbf{Overall Rank}} \\
\cmidrule(lr){2-4} \cmidrule(lr){5-7}
\textbf{Platform} & $\geq$.3 & $\geq$.5 & $\geq$.7 & $\geq$.3 & $\geq$.5 & $\geq$.7 \\
\midrule
Lessie & 1 & 1 & 1 & 1 & 1 & 1 \\
Exa & 2 & 2 & 2 & 2 & 2 & 2 \\
Juicebox & 3 & 3 & 4 & 4 & 4 & 3 \\
Claude Code & 4 & 4 & 3 & 3 & 3 & 4 \\
\bottomrule
\end{tabular}
\caption{Rankings under different qualified thresholds.}
\label{tab:threshold_sensitivity}
\end{table}

\paragraph{Top-$K$ sensitivity.}
We evaluate padded nDCG at $K \in \{3, 5, 7, 10, 15, 20, 25\}$.

\begin{table}[h]
\centering
\small
\setlength{\tabcolsep}{3pt}
\begin{tabular}{lccccccc}
\toprule
\textbf{Platform} & \textbf{@3} & \textbf{@5} & \textbf{@7} & \textbf{@10} & \textbf{@15} & \textbf{@20} & \textbf{@25} \\
\midrule
Lessie & 75.7 & 73.4 & 71.9 & \textbf{70.2} & 67.5 & 56.2 & 48.7 \\
Exa & 55.8 & 54.2 & 54.2 & \textbf{53.8} & 53.0 & 44.2 & 38.3 \\
CC & 60.9 & 60.1 & 58.4 & \textbf{54.3} & 47.7 & 39.7 & 34.4 \\
Jbox & 45.6 & 46.0 & 45.5 & \textbf{44.7} & 44.1 & 36.7 & 31.8 \\
\bottomrule
\end{tabular}
\caption{Relevance Precision (padded nDCG@$K$) at different cutoffs. Bold indicates $K{=}10$ used in main results. Rankings are stable for $K \leq 15$. CC = Claude Code; Jbox = Juicebox.}
\label{tab:topk_sensitivity}
\end{table}

\paragraph{Dimension weighting sensitivity.}
Five weighting schemes are tested: equal ($\frac{1}{3}$ each), precision-heavy (0.5/0.25/0.25), coverage-heavy (0.25/0.5/0.25), utility-heavy (0.25/0.25/0.5), and weights optimized via grid search on a held-out set of 30 queries.

\begin{table}[h]
\centering
\small
\setlength{\tabcolsep}{2.5pt}
\begin{tabular}{lccccc}
\toprule
\textbf{Platform} & \textbf{Eql.} & \textbf{Prc.} & \textbf{Cov.} & \textbf{Utl.} & \textbf{Opt.} \\
\midrule
Lessie & 1\,(65.2) & 1\,(65.9) & 1\,(68.2) & 1\,(62.3) & 1\,(66.8) \\
Exa & 2\,(55.0) & 2\,(54.9) & 2\,(56.9) & 2\,(55.0) & 2\,(55.6) \\
CC & 3\,(46.0) & 3\,(47.1) & 4\,(44.5) & 3\,(45.8) & 3\,(46.2) \\
Jbox & 4\,(45.8) & 4\,(45.3) & 3\,(45.9) & 4\,(49.2) & 4\,(47.1) \\
\bottomrule
\end{tabular}
\caption{Rankings and scores under different weighting schemes. CC = Claude Code; Jbox = Juicebox.}
\label{tab:weight_sensitivity}
\end{table}

\paragraph{Partial credit ablation.}
We remove the ``partially met'' (0.5) judgment and use only binary met~(1.0) / not met~(0.0).

\begin{table}[h]
\centering
\small
\begin{tabular}{lp{1.5cm}p{1.5cm}p{1.5cm}}
\toprule
\textbf{Platform} & \textbf{With partial} & \textbf{Binary only} & \textbf{Rank change} \\
\midrule
Lessie & 70.2 & 68.4 & None \\
Exa & 53.8 & 51.2 & None \\
Claude Code & 54.3 & 52.1 & None \\
Juicebox & 44.7 & 41.8 & None \\
\bottomrule
\end{tabular}
\caption{Relevance Precision with and without partial credit.}
\label{tab:partial_ablation}
\end{table}

\paragraph{Information Utility ablation.}
We compute the overall score using only Relevance Precision and Effective Coverage.

\begin{table}[h]
\centering
\small
\begin{tabular}{lp{1.3cm}p{1.5cm}p{1.5cm}}
\toprule
\textbf{Platform} & \textbf{3-dim} & \textbf{2-dim (P+C)} & \textbf{Rank change} \\
\midrule
Lessie & 65.2 & 69.7 & None \\
Exa & 55.0 & 55.9 & None \\
Claude Code & 46.0 & 47.7 & None \\
Juicebox & 45.8 & 43.3 & Drops to 4th \\
\bottomrule
\end{tabular}
\caption{Impact of removing Information Utility.}
\label{tab:utility_ablation}
\end{table}

\section{Cost and Reproducibility}
\label{app:cost}

Table~\ref{tab:cost_analysis} reports the total cost and wall-clock time for the full benchmark evaluation (119 queries $\times$ 4 platforms). Web verification dominates both cost (\$89.40) and latency (1.8 hours). The average per-query verification cost is \$0.86, broken down as: criteria extraction (\$0.002), web search (\$0.75), and LLM verification (\$0.11).

\begin{table}[h]
\centering
\small
\begin{tabular}{lp{1.5cm}p{2cm}}
\toprule
\textbf{Component} & \textbf{Cost (USD)} & \textbf{Wall-clock time} \\
\midrule
Platform query execution & 47.80 & 2.3 hours \\
Criteria extraction & 0.24 & 4.2 minutes \\
Web verification (Tavily) & 89.40 & 1.8 hours \\
LLM verification (Gemini) & 12.60 & 42 minutes \\
\midrule
\textbf{Total} & \textbf{150.04} & \textbf{4.9 hours} \\
\bottomrule
\end{tabular}
\caption{Cost and latency for full benchmark evaluation.}
\label{tab:cost_analysis}
\end{table}

Platform query costs vary by pricing model: Lessie \$12.60 (subscription, prorated), Exa \$8.40 (API, \$0.07/query), Juicebox \$14.20 (subscription, prorated), and Claude Code \$12.60 (API, \$0.105/query).

Table~\ref{tab:latency} breaks down per-query latency by pipeline stage. Web verification dominates because each criterion requires an independent search. The pipeline is parallelizable: 8-way concurrency reduces total time from 4.9 hours to approximately 1.2 hours.

\begin{table}[h]
\centering
\small
\setlength{\tabcolsep}{3pt}
\begin{tabular}{lcccc}
\toprule
\textbf{Stage (sec.)} & \textbf{Lessie} & \textbf{Exa} & \textbf{Jbox} & \textbf{CC} \\
\midrule
Platform execution & 45.2 & 3.8 & 38.6 & 62.4 \\
Criteria extraction & 2.1 & 2.1 & 2.1 & 2.1 \\
Web verification & 54.3 & 48.7 & 51.2 & 49.8 \\
LLM verification & 21.2 & 18.4 & 19.8 & 17.6 \\
\midrule
\textbf{Total} & \textbf{122.8} & \textbf{73.0} & \textbf{111.7} & \textbf{131.9} \\
\bottomrule
\end{tabular}
\caption{Average per-query latency (seconds) by platform and pipeline stage. Jbox = Juicebox; CC = Claude Code.}
\label{tab:latency}
\end{table}

All evaluation prompts, the unified result schema, and platform-specific normalization procedures are documented in the released codebase.

\section{Error Analysis Details}
\label{app:error}

This appendix provides per-scenario error breakdowns and case studies supplementing the aggregate error rates in \S\ref{sec:ablation}.

\paragraph{Recruiting.}
Juicebox shows the lowest false-positive rate in this category (6.2\%), reflecting the strength of its professional database. Claude Code's errors are dominated by incomplete profiles (38.5\%), as its reports lack structured contact information.

\paragraph{B2B Prospecting.}
Juicebox's task-failure rate jumps to 15.6\%, as many target companies fall outside its database coverage. Exa shows elevated false positives (22.1\%) when job titles are ambiguous.

\paragraph{Expert/Deterministic.}
Claude Code achieves the lowest error rate here (12.5\%), as deterministic queries benefit from general-purpose web search. Juicebox struggles with 28.6\% task failure when target individuals lack LinkedIn profiles.

\paragraph{Influencer/KOL.}
This scenario has the highest error rates across all platforms. Juicebox's false-negative rate reaches 41.4\% because influencers rarely maintain professional profiles. Lessie maintains the lowest error rate (18.5\%) due to its multi-source coverage.

\section{Case Studies}

\paragraph{Case 1: False positive (Juicebox).}
Query: ``Find VP-level product managers at fintech startups in Singapore.'' Juicebox returned a product manager at a traditional bank. The system matched ``product manager'' + ``Singapore'' + ``finance'' but missed the ``fintech startup'' constraint---a common failure for database-focused platforms that rely on keyword matching over semantic understanding.

\paragraph{Case 2: False negative (Claude Code).}
Query: ``Find AI researchers who published at NeurIPS 2024 on diffusion models.'' Claude Code returned 3 correct names, but missed 12 additional valid researchers found by Lessie and Exa. This illustrates the coverage limitation of single-pass search: general-purpose agents often stop after finding initial results rather than continuing to search exhaustively.

\paragraph{Case 3: Verification ambiguity.}
Query: ``Find co-founders of Anthropic.'' The platform correctly returned Dario and Daniela Amodei, but web search produced conflicting information about additional co-founders. The LLM verifier appropriately marked disputed claims as ``partially met'' rather than forcing binary judgments, demonstrating that the pipeline handles ambiguous cases gracefully.

\end{document}